\documentclass[journal,12pt,onecolumn,draftclsnofoot]{IEEEtran}
\usepackage{amsmath,amsfonts}
\usepackage{amssymb}
\usepackage{algorithmic}
\usepackage{algorithm}
\usepackage{array}
\usepackage{textcomp}
\usepackage{stfloats}
\usepackage{url}
\usepackage{verbatim}
\usepackage{graphicx}
\usepackage{cite}
\newcommand{\Luka}{\L{}ukasiewicz }
\usepackage{xcolor}
\usepackage{soul}
\usepackage{subcaption}
\usepackage{tikz}
\newcommand\figref{Fig.~\ref}
\newcommand\defref{Definition~\ref}
\newcommand\lemref{Lemma~\ref}
\newcommand\thmref{Theorem~\ref}
\newcommand\secref{Section~\ref}
\newcommand\appref{Appendix~\ref}
\newtheorem{theorem}{Theorem}[section]
\newtheorem{lemma}[theorem]{Lemma}

\newtheorem{definition}[theorem]{Definition}
\newcommand{\Acal}{\mathcal{A}}

\begin{document}
\setstcolor{blue}
\title{Extracting Formulae in Many-Valued Logic from Deep Neural Networks}

\author{Yani Zhang and Helmut Bölcskei \thanks{The authors are with the Chair for Mathematical Information Science, ETH Zurich, Switzerland. H.~B\"olcskei gratefully acknowledges support by the Lagrange Mathematics and Computing Research Center, Paris, France.} }



\maketitle

\begin{abstract}
We propose a new perspective on deep rectified linear unit (ReLU) networks, namely as circuit counterparts of \Luka infinite-valued logic---a many-valued (MV) generalization of Boolean logic. An algorithm\footnote{A Python implementation for ReLU networks with 
integer, rational, or real weights is available at  
\url{https://www.mins.ee.ethz.ch/research/downloads/NN2MV.html}} for extracting formulae in MV logic from trained deep ReLU networks is presented. The algorithm respects the network architecture, in particular compositionality, 
thereby honoring algebraic information present in the training data. 
We also establish the representation benefits of deep networks from a mathematical logic perspective.
\end{abstract}


\section{Introduction and Contributions}

\IEEEPARstart{S}{tate-of-the-art} deep neural networks often exhibit remarkable reasoning capabilities, e.g., in mathematical tasks~\cite{lample2019deep}, program synthesis~\cite{bubeck2023sparks}, and algorithmic reasoning~\cite{weiss2021thinking}.
In an attempt to develop an understanding of this phenomenon, the present paper establishes a systematic connection between neural networks and mathematical logic. Specifically, we shall be interested in reading out logical formulae from (trained) deep neural networks, with the underlying idea that these formulae yield a logical description of the data the network was trained on. 

Consider a neural network that realizes a map $f: [0,1]^n \rightarrow [0,1]$. Taking a step back, we note that for input variables and the output 
taking on the values $0$ and~$1$ only,~$f$ reduces to a Boolean function and can hence be studied by means of Boolean algebra, see e.g.,~\cite{odonnell-boolean}. 
The algebraic expression of this Boolean function---in terms of AND, OR, and NOT operations---provides a logical description of the underlying functional relationship. 
Here, we shall be concerned with the generalization of 
this correspondence from Boolean functions~$f:\{0,1\}^n \rightarrow \{0,1\}$ to general functions~$f:[0,1]^n\rightarrow [0,1]$. 
This immediately leads to the following questions:
\begin{enumerate}
    \item What is the logical system replacing Boolean logic? 
    \item How can formulae in this logical system be extracted from neural networks realizing $f: [0,1]^n \rightarrow [0,1]$?
\end{enumerate}
As for the first question, we shall show that the theory of infinite-valued \Luka logic~\cite{cignoli2013algebraic}, sometimes also referred to as \L{}ukasiewicz-Tarski logic, provides a suitable framework for characterizing general functions $f: [0,1]^n \rightarrow [0,1]$ from a logical perspective. With slight abuse of terminology, we shall refer to infinite-valued \Luka logic as many-valued~(MV) logic throughout the paper. Based on a fundamental result~\cite{mcnaughton1951theorem}, which characterizes the class of truth functions in MV logic---also called McNaughton functions---as continuous piecewise linear functions with integer coefficients, we show that neural networks with the ReLU nonlinearity~$\rho(x) = \max\{0, x\}$ and integer weights\footnote{By weights, we mean the entries of the weight matrices and bias vectors associated with the network.} naturally implement statements in MV logic and vice versa. 
We develop an algorithm for extracting formulae in MV logic from McNaughton functions specified through deep ReLU networks. 
There are two procedures in the literature for extracting MV formulae from McNaughton functions, namely the Schauder hat method \cite{mundici1994constructive} and the hyperplane method \cite{Aguzollio1998thesis}, both developed in the course of alternative and constructive proofs of the McNaughton theorem~\cite{mcnaughton1951theorem} and
applying to the one-dimensional case $n=1$ only, while our algorithm works for arbitrary dimension~$n$. 
Moreover, for $n=1$, by virtue of honoring the compositional structure of deep ReLU networks, 
the algorithm we propose can result in significantly shorter formulae than~\cite{mundici1994constructive} and~\cite{Aguzollio1998thesis}. This will be established analytically and illustrated through numerical results. 


In practice, trained neural networks will not exhibit integer weights, unless explicitly enforced in the training process. Extensions of MV logic, namely Rational \Luka logic~\cite{gerla2001rational} and~$\mathbb{R}\mathcal{L}$~\cite{di2014lukasiewicz}, have truth functions that are again continuous piecewise linear, but with rational and real coefficients, respectively. Such functions are likewise naturally realized by ReLU networks, but correspondingly with rational and real weights.
For pedagogical reasons and to render the presentation more accessible, we first present the entire program outlined above, including the algorithm for extracting MV formulae from ReLU networks, for the case of integer weights and then describe the extensions to the rational and real
cases. 

The ideas presented in this paper are inspired by~\cite{1018111, amato2005neural, di2013adding}. Specifically, Amato et al.~\cite{1018111, amato2005neural} pointed out that neural networks, with the clipped ReLU (CReLU) nonlinearity~$\sigma(x) = \min\{1,\max\{0,x\}\}$ and rational weights, realize truth functions in Rational \Luka logic. Di Nola et al.~\cite{di2013adding} proved that CReLU networks with real weights realize truth functions in~$\mathbb{R}\mathcal{L}$ logic. The universal correspondence between ReLU networks and MV logic, Rational \Luka logic, and~$\mathbb{R}\mathcal{L}$ along with the algorithm for extracting logical formulae from ReLU networks appear to be entirely new. 


\section{Boolean Logic and MV Logic}\label{sec:basiclogic}
We start with a brief review of the basic concepts in Boolean and MV logic. Boolean algebra on the set~$\{0,1\}$ consists of the application of the logical operations \texttt{AND}, \texttt{OR}, and \texttt{NOT}, denoted by~$\odot, \oplus$, and~$\lnot$, respectively, to Boolean propositional variables taking value in $\{0,1\}$. The algebra is fully characterized through the following identities on Boolean variables $x,y,z$:
\begin{equation}\label{eq:BooleanAxm}
\begin{aligned}[c]
& x \oplus 0 =x \hspace{1.6cm}  x \odot 1 = x \\
& x \oplus \lnot x = 1 \hspace{1.4cm}  x \odot \lnot x= 0\\
& x\oplus y = y\oplus x \hspace{1cm}  x\odot y = y\odot x \\
& (x \oplus y) \oplus z = x \oplus (y\oplus z) \\
&  (x \odot y) \odot z = x \odot (y\odot z) \\
 &   x\oplus (y \odot z) = (x\oplus y)\odot (x\oplus z)\\
 &   x\odot(y\oplus z) = (x\odot y) \oplus (x\odot z)
\end{aligned}
\end{equation}
Note that one can define the operation $\odot$ in terms of $\oplus$ and~$\lnot$ according to $x\odot y:= \lnot (\lnot x \oplus \lnot y)$.

MV logic~\cite{tarski1983logic} generalizes Boolean logic by allowing for propositional variables that take truth values in the interval~$[0,1]$. 
The corresponding algebraic counterpart is known as Chang's MV algebra~\cite{chang1958algebraic} (see \defref{def:mv-algebra}). 
We proceed to the definition of the so-called standard MV algebra.

\begin{definition}\label{def:MV01} 
     Consider the unit interval~$[0, 1]$, and define~$x \oplus y = \min \{1, x+y\}$ and~$\lnot x = 1-x$, for~$x, y \in [0,1]$. It can be verified that the structure~$\mathcal{I} = \langle [0,1], \oplus, \lnot,0 \rangle$ is an MV algebra~\cite{cignoli2013algebraic}. In particular,~$\mathcal{I}$ constitutes the algebraic counterpart of \Luka infinite-valued logic~\cite{chang1958algebraic}. We further define the operation~$x  \odot y := \lnot(\lnot x \oplus \lnot y) = \max\{0,x+y-1\}$.
\end{definition}

It can be shown that the Boolean algebra $\mathcal{B}:= \langle \{0,1\}, \texttt{OR}, \texttt{NOT}, 0\rangle$ is a special case of MV algebras. The MV algebra $\mathcal{I}$ in Definition \ref{def:MV01} 
is referred to as standard because an equation holds in every MV algebra iff it holds in~$\mathcal{I}$~\cite{chang1958algebraic, chang1959new}. Additional relevant material on MV algebras is provided in~\appref{app:MValge}.

\section{Extraction of formulae in MV logic from ReLU networks}\label{sec:extract}
MV terms (see Definition~\ref{def:mv-term}) are finite strings composed of propositional variables $x_1,x_2,\ldots\,$ connected by~$\oplus$, $\odot$, and~$\lnot$ operations and brackets $( )$, such as e.g.,~$(x_1\oplus \lnot x_2) \odot x_3$. We define the length of an MV term as the total number of occurrences of propositional variables, e.g., $(x_1\odot \lnot x_1) \oplus x_2$ is of length $3$.
Term functions (see Definition~\ref{def:term-function}) are the corresponding truth functions obtained by interpreting the logical operations according to how they are specified in the concrete MV algebra used, e.g., $x \oplus y = \min \{1, x+y\}$ in the standard MV algebra $\mathcal{I}$.
In the Boolean algebra $\mathcal{B}$, term functions are binary tables~$\{ \{0,1\}^n \rightarrow \{0,1\}:n\in \mathbb{N} \}$. In the standard MV algebra $\mathcal{I}$, term functions are characterized by continuous piecewise linear functions with integer coefficients as formalized by the McNaughton theorem~\cite{mcnaughton1951theorem}.

\begin{theorem}[\!\!\cite{mcnaughton1951theorem}]\label{them:McNaughtonTheorem}
Consider the MV algebra~$\mathcal{I}$ in~\defref{def:MV01}. Let~$n\in \mathbb{N}$. For a function~$f: [0,1]^n \rightarrow [0,1]$ to have a corresponding MV term~$\tau$ such that the associated term function~$\tau ^{\mathcal{I}}$ satisfies~$\tau ^{\mathcal{I}} = f$ on~$[0,1]^n$, it is necessary and sufficient that~$f$ satisfy the following conditions: 
\begin{enumerate}
\item $f$ is continuous with respect to the natural topology on~$[0,1]^n$,
    \item there exist linear polynomials~$p_1, \ldots \hspace{-0.02cm}, p_\ell$ with integer coefficients, i.e., 
    \begin{equation*}
        p_j(x_1, \ldots \hspace{-0.02cm}, x_{n}) = m_{j1}x_1+\cdots+m_{jn}x_{n}+b_j, 
    \end{equation*} for~$j = 1,\ldots,\ell,$ with~$ m_{j1},\ldots \hspace{-0.02cm}, m_{jn},b_j \in \mathbb{Z}$, such that for every~$x \in [0,1]^n$, there is a~$j \in \{1,\ldots \hspace{-0.02cm}, \ell\}$ with~$ f(x) = p_j(x)$.
\end{enumerate} Functions satisfying these conditions are called McNaughton functions. 
\end{theorem}

ReLU networks (see~\defref{def:ReLUNN}) are compositions of affine transformations and the ReLU nonlinearity~$\rho (x)= \max\{0,x\}$ (applied element-wise) and as such realize continuous piecewise linear functions. Specifically, the class of ReLU networks with integer weights is equivalent to the class of formulae in MV logic. The corresponding formal statement is as follows.

\begin{theorem}\label{them:NN-MV}
For~$n\in \mathbb{N}$, let~$\tau(x_1,\ldots, x_n)$ be an MV term in~$n$ variables with~$\tau^{\mathcal{I}}:[0,1]^n \rightarrow [0,1]$ the associated term function in~$\mathcal{I}$. There exists a ReLU network~$\Phi$ with integer weights, satisfying 
\[
 \Phi(x_1,\ldots, x_n)=\tau^{\mathcal{I}} (x_1,\ldots,x_n), 
\]
for all $(x_1,\ldots,x_n) \in [0,1]^n$. Conversely, for every ReLU network $\Phi  : [0,1]^n\rightarrow [0,1]$ with integer weights, there exists an MV term $\tau(x_1,\ldots,x_n)$ whose associated term function in $\mathcal{I}$ satisfies
\[
\tau^{\mathcal{I}} (x_1,\ldots,x_n) = \Phi(x_1,\ldots, x_n), 
\]
for all $(x_1,\ldots,x_n) \in [0,1]^n$.
\end{theorem}

The remainder of this section is devoted to the proof of~\thmref{them:NN-MV}, along with the development of an algorithm for extracting formulae in MV logic from ReLU networks with integer weights. 

First, we show how, for a given MV term~$\tau$, a ReLU network with integer weights realizing the associated term function~$\tau^{\mathcal{I}}$ can be constructed. 
To this end, we note that the operation $\lnot x = 1-x$, by virtue of being affine, is trivially realized by a ReLU network. Further, there exist ReLU networks $\Phi^{\oplus}$ and $\Phi^{\odot }$, with integer weights, realizing the $\oplus$ and $\odot$ operations in $\mathcal{I}$, i.e.,
\begin{align*}
\label{eq:networkAND}    \Phi^{\oplus }(x,y) &= \min\{1,x+y\} \\
    \Phi^{\odot }(x,y) &= \max\{0,x+y-1\},
\end{align*} for all $x,y\in [0,1]$. Detailed constructions of $\Phi^\oplus$ and $\Phi^\odot$ are provided in~\lemref{lem:NNminmax}. 
According to \lemref{lem:ReLUConcatenating}~\cite{elbrachter2021deep}, compositions of ReLU networks are again ReLU networks. The ReLU network realizing the term function associated with the MV term~$\tau$ can hence be obtained by concatenating ReLU networks implementing the operations $\oplus, \odot,$ and $\lnot$ as they appear in the expression for $\tau$. What is more, inspection of the proof of~\lemref{lem:ReLUConcatenating} reveals that the integer-valued nature of the weights is preserved under composition. Therefore, the resulting overall ReLU network has integer weights. 
We illustrate this procedure by way of the simple example 
$\tau = (x \oplus x) \odot \lnot y$, which yields the associated ReLU network (see \appref{app:tau1} for details)
\begin{equation}\label{eq:tau}
 \Phi^\tau = W_3 \circ \rho \circ W_2 \circ \rho \circ W_1,   
\end{equation}
where 
\begin{equation*}
\begin{aligned}
    W_1(x,y) &= \begin{pmatrix}
    -2 & 0 \\
    0 & 1\\
    0 & -1
\end{pmatrix}\begin{pmatrix}
    x \\y
\end{pmatrix} + \begin{pmatrix}
    1 \\ 0 \\ 0
\end{pmatrix}\hspace{-0.1cm},\quad x,y \in \mathbb{R}, \\
W_2(x) &= \begin{pmatrix}
   -1 & -1 & 1
\end{pmatrix}x+1, \quad x\in \mathbb{R}^3,\\
W_3(x) &=  x , \quad x\in \mathbb{R}.
\end{aligned}    
\end{equation*}

The proof of the converse statement in~\thmref{them:NN-MV} will be effected in a constructive manner, in the process developing an algorithm for extracting the MV term encoded by a given ReLU network with integer weights. 
For convenience of exposition, henceforth we apply $\rho$ to the output of ReLU networks (see Definition~\ref{def:ReLUNN}), e.g., we write $\rho \circ W_2 \circ \rho \circ W_1$ instead of $W_2 \circ \rho \circ W_1$; this does not affect
the network input-output relation as we only consider networks realizing McNaughton functions and McNaughton functions always map to $[0,1]$.

Honoring the compositional structure of deep ReLU networks, our algorithm proceeds in a compositional manner. Specifically, it consists of the following three steps.

\textit{Step 1}: Transform the ReLU network (with integer weights) into an equivalent network employing the CReLU nonlinearity $\sigma(x) = \min\{1, \max\{0,x \}\}$. This is done by exploiting the fact that the domain of the ReLU network is the unit cube~$[0,1]^n$ and, consequently, with finite-valued weights, the input to all neurons in the network is bounded. Concretely, if the input of a given $\rho$-neuron is contained in the interval $[-A,B]$, with $A,B \in \mathbb{R}_{\geqslant 0}$ and $B\geq A$, we replace this $\rho$-neuron by one or multiple $\sigma$-neurons (shifted by integer values) according to 
\begin{equation}\label{eq:sigma-replacement}
\begin{aligned}
     \rho(t) &= 
     \begin{cases}
    \sigma(t), & B \in (0,1]\\
        \sigma(t) + \sigma(t-1), & B \in (1,2]\\
        \sigma(t)+\sigma(t-1)+\cdots+\sigma(t-\lfloor B \rfloor), & B > 2.
        \end{cases} 
\end{aligned}    
\end{equation}
Neurons that have $B=0$ are deleted. Applying these replacements to all neurons in the $\rho$-network, we obtain a $\sigma$-network with integer weights and the same input-output relation as the $\rho$-network.



\textit{Step 2}: Extract MV terms from individual $\sigma$-neurons, which are of the form 
\begin{equation}\label{eq:sigmaneuron}
    \sigma(m_1x_1+\cdots+m_nx_n+b),
\end{equation} with $m_1,\ldots,m_n, b\in \mathbb{Z}$. The following lemma, a proof of which can be found in Appendix~\ref{app:proofLem4.4}, forms the basis for accomplishing this in an iterative manner.
\begin{lemma}[\!\!\cite{rose1958, mundici1994constructive}]\label{lem:extractmv}
Consider the function $f(x_1,\ldots, x_n) = m_1x_1+\cdots+m_nx_n+b,$ $(x_1,\ldots,x_n) \in [0,1]^n$, with $ m_1,\ldots \hspace{-0.02cm},m_n, b\in \mathbb{Z}.$ W.l.o.g. assume that $\max_{i=1}^n |m_i| = m_1$. Let $f_{\circ}(x_1,\ldots, x_n) =  (m_1-1)x_1+m_2x_2+\cdots+m_nx_n+b$. Then,
    \begin{equation}\label{eq:lemma4.4}
        \sigma(f) = (\sigma(f_{\circ}) \oplus x_1) \odot \sigma(f_{\circ}+1).
    \end{equation}
\end{lemma} 

Before proceeding to the next step, we demonstrate the iterative application of \lemref{lem:extractmv} to the example $\sigma$-neuron $\sigma(x_1-x_2+x_3-1)$. First, we eliminate the variable $x_1$ according to
\begin{equation}\label{eq:sigma}
     \sigma(x_1-x_2+x_3-1)  =  (\sigma( -x_2+x_3-1 ) \oplus x_1) \odot \sigma( -x_2+x_3 ).
\end{equation}
Next, we remove $x_3$ inside $\sigma(-x_2+x_3-1)$ and $\sigma( -x_2+x_3 )$ according to
\begin{align}
\label{eq:sigma1} \sigma( -x_2+x_3-1 ) &= (\sigma(-x_2 -1) \oplus x_3) \odot     \sigma(-x_2) \\
\label{eq:sigma2} \sigma( -x_2+x_3 ) &= (\sigma(-x_2 ) \oplus x_3) \odot     \sigma(-x_2+1). 
\end{align}
We then note that $\sigma(-x_2)=0$ as $x_2 \in [0,1]$, and owing to $x\odot 0 = 0$, for $x\in [0,1]$,~\eqref{eq:sigma1} reduces to $ \sigma( -x_2+x_3-1 )=0$. Likewise, in~\eqref{eq:sigma2} we get $\sigma(-x_2) \oplus x_3 = x_3$. We can now further simplify~\eqref{eq:sigma2} according to
\begin{align}
 \label{eq:sigma3}   x_3\odot \sigma(-x_2+1) &=x_3\odot ( 1-\sigma(x_2) )\\
 \label{eq:sigma4}   &= x_3\odot \lnot x_2,
\end{align}
where in~\eqref{eq:sigma3} we used $\sigma(x) = 1-\sigma(-x+1)$, for $x\in \mathbb{R}$, and~\eqref{eq:sigma4} is by $\sigma(x) = x$ and $\lnot x=1-x$, both for $x\in [0,1]$. Substituting the simplified results of~\eqref{eq:sigma1} and~\eqref{eq:sigma2} back into~\eqref{eq:sigma}, we obtain the MV term corresponding to $ \sigma(x_1-x_2+x_3-1)$ as $x_1\odot (x_3\odot \lnot x_2)$. 

For later use, we also state a result on the length of MV terms 
obtained through iterative application of Lemma~\ref{lem:extractmv}.
\begin{lemma}\label{lem:length_sigma}
Consider $f(x_1,\ldots, x_n) = m_1x_1+\cdots+m_nx_n+b,$ $(x_1,\ldots,x_n) \in [0,1]^n$, with $ m_1,\ldots \hspace{-0.02cm},m_n, b\in \mathbb{Z}$.
The MV term corresponding to the function $\sigma( f)$, obtained by iteratively applying Lemma \ref{lem:extractmv} has length at most $2^m-1$, where
$m := \sum_{i=1}^{n} |m_i|$.
\end{lemma}
The proof of Lemma \ref{lem:length_sigma} can be found in Appendix \ref{app:proof_length_sigma}.




\textit{Step 3}: Compose the MV terms corresponding to the individual $\sigma$-neurons according to the layered structure of the CReLU network to get the MV term associated with the ReLU network. To illustrate this step, suppose that the neurons $\sigma^{(1)}(\cdot)$ and $\sigma^{(2)}(\cdot)$ have associated MV terms~$\tau^{(1)}$ and~$\tau^{(2)}$, respectively, and a third neuron $\sigma^{(3)}(m_1x_1+m_2x_2+b)$ has associated MV term~$\tau^{(3)}(x_1,x_2)$. The MV term corresponding to the CReLU network $\sigma^{(3)}(m_1\sigma^{(1)}+m_2\sigma^{(2)}+b)$ is obtained by replacing all occurrences of~$x_1$ in~$\tau^{(3)}$ by~$\tau^{(1)}$ and all occurrences of~$x_2$ by~$\tau^{(2)}$. This finalizes the proof of \thmref{them:NN-MV}.


The essence of the proof of \thmref{them:NN-MV} resides in a strong algebraic property shared by the standard MV algebra $\mathcal{I}$ and ReLU networks. Concretely, compositions of ReLU networks with integer weights again yield ReLU networks with integer weights, and compositions of formulae in MV logic result in formulae in MV logic. As we shall see in~\secref{sec:4}, this parallelism 
extends to the cases of ReLU networks with rational weights and Rational \Luka logic as well as ReLU networks with real weights and $\mathbb{R}\mathcal{L}$.

\section{Existing results on MV term extraction from McNaughton functions}\label{sec:review}
Recalling that a ReLU network with integer weights realizes a continuous piecewise linear function with integer coefficients, the algorithm devised in the previous section can equivalently be seen as extracting an MV term from the McNaughton function realized by the network. 
As mentioned in the introduction, there are two algorithms in the literature for extracting MV terms from McNaughton functions, namely the Schauder hat method \cite{mundici1994constructive} and the hyperplane method \cite{Aguzollio1998thesis}, both of which apply, however, only to the one-dimensional case $n=1$. 
The basic tenets of these two algorithms can be summarized as follows.

The Schauder hat method \cite{mundici1994constructive} constructs so-called Schauder hats, i.e., functions of pyramidal shape, supported on unions of simplices. More concretely, 
a simplicial complex over $[0,1]$ obtained by splitting the unit cube according to different permutations of the linear pieces of the McNaughton function $f$ is subdivided into a unimodular 
simplicial complex. Thanks to unimodularity, each Schauder hat can then be expressed in terms of ``$\min$'' and ``$\max$'' operations, which are, in turn, realizable by the operations $\oplus, \odot$, and $\lnot$ according to
\begin{equation}\label{eq:minmax}
\begin{aligned}
 \min\{x,y\}&=\lnot (\lnot x \odot y)\odot y := x\wedge y\\
    \max\{x,y\}&=\lnot (\lnot x \oplus y)\oplus y := x\vee y.
\end{aligned}    
\end{equation}
The overall MV term corresponding to $f$ is finally obtained by combining the MV terms associated with the individual Schauder hats through $\oplus$ operations. 
We refer the reader to~\cite{mundici1994constructive} for a detailed account of the algorithm. 

The hyperplane method~\cite{Aguzollio1998thesis} expresses the McNaughton function $f$, with linear pieces~$p_1,\ldots, p_\ell$, in terms of the truncated linear polynomials $\sigma(p_1), \ldots, \sigma(p_\ell)$, where $\sigma(x)= \min\{1,\max\{0,x\}\}$, for $x\in \mathbb{R}$, according to
\begin{equation}\label{eq:A}
   f = \min_I \max_J \sigma(p_i), 
\end{equation} 
where $I, J \subset \{1,\ldots,\ell\}$ are index sets. Next, the MV terms corresponding to $\sigma(p_1),\ldots, \sigma(p_\ell)$ are determined by repeated application of~\lemref{lem:extractmv}. These MV terms are finally combined into the MV term corresponding to $f$ by using \eqref{eq:A} and \eqref{eq:minmax}. 

The only result on MV term extraction in the multi-dimensional case we are aware of is \cite[Theorem~3]{mundici1994constructive}, which establishes
the existence of a Turing machine delivering the MV term underlying a given McNaughton function. 

\section{Deep networks yield shorter formulae}\label{sec:comparision}

\begin{figure}[tb]
\begin{center}
\centerline{\includegraphics[width=0.55\columnwidth]{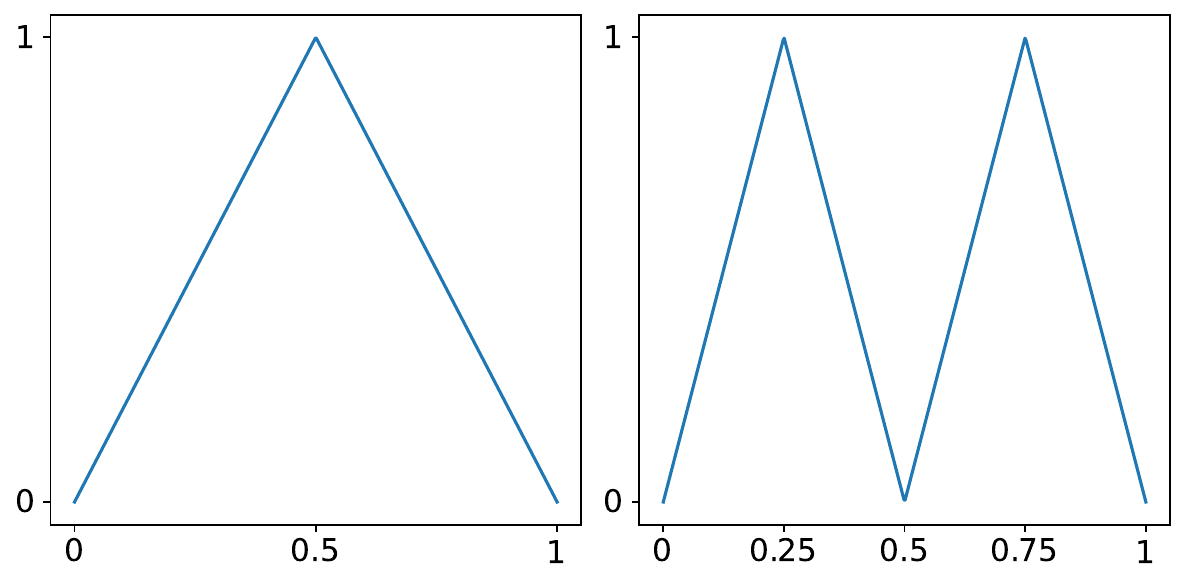}}
\caption{Left: The function $g$, right: The function $g_2$.}
\label{fig:hat}
\end{center}
\vskip -0.2in
\end{figure}

In this section, we perform an in-depth comparison between our algorithm for the univariate case, the Schauder hat method, and the hyperplane method.
Specifically, we shall be 
interested in the length of the logical formulae produced by the different algorithms. We will establish, in the case of our method, that deep networks tend to produce shorter formulae.
For illustration purposes, we let our discussion be guided by the following example. 
Consider the hat function $g:[0,1] \rightarrow [0,1]$ in \figref{fig:hat},
\begin{equation}\label{eq:g}
\begin{aligned}
     g(x) &= \rho(2x)-2\rho(2x-1)  \\
  &= \begin{cases}
    2x, &0\leq x\leq \frac{1}{2}\\
    2-2x, &\frac{1}{2} < x\leq 1.
\end{cases} 
\end{aligned}    
\end{equation}
Let $g_1(x) = g(x)$, and define the $s$-th order sawtooth function $g_s$ as the $s$-fold composition of $g$ with itself, i.e.,
\begin{equation}\label{eq:sawtooth}
   g_s:= \underbrace{g\circ \cdots \circ g}_{s}, \quad  s\geq 2. 
\end{equation}


\noindent We note that $g_s$ consists of the $2^s$ linear regions
\[
\left[0,\frac{1}{2^s}\right], \left[\frac{1}{2^s}, \frac{2}{2^s}\right], \ldots, \left[\frac{2^s-1}{2^s}, 1\right],
\]
with slope on each of these regions given by either $2^s$ or $-2^s$. As all the coefficients of $g_s$ are integer, it constitutes a valid McNaughton function.


\subsection{Analytical characterization of MV term lengths}\label{subsec:analysis}

We start by analytically characterizing the length of the MV terms corresponding to $g_s$ produced by the different algorithms.

{\em The Schauder hat method.} 
Permutations of the linear pieces of $g_s$ subdivide the interval $[0,1]$ into $2^s$ simplices, namely $[0,\frac{1}{2^s}], [\frac{1}{2^s}, \frac{2}{2^s}], \ldots, [\frac{2^s-1}{2^s}, 1]$. For $s \geq 2$, these simplices are further subdivided to ensure the unimodularity condition. The number of Schauder hats is\footnote{We write $f = \Omega (g)$ to signify that $\limsup_{t\rightarrow \infty}\left|\frac{f(t)}{g(t)} \right| > 0 $.} $\Omega(2^s)$. As the MV term 
associated with each Schauder hat is of length at least $2$, the overall MV term corresponding to $g_s$ has length at least 
$\Omega(2^{s+1})$. As illustrated by the numerical results in Fig.~\ref{fig:res}, the actual length grows (as a function of $s$) significantly faster than this loose lower bound suggests.

{\em The hyperplane method.}
The MV term associated with $g_s$ obtained from the hyperplane method is of the form 
\begin{equation}\label{eq:hyperplane_norm}
    \wedge_{I} \vee_{J} \sigma(p_i), 
\end{equation}
where $I$ and $J$ are index sets. 
The cardinality of $I$ is equal to the number of linear pieces of 
$g_s$, which is $2^s$ and the cardinality of $J$ is at least one (see \cite[Sec. 1.4]{Aguzollio1998thesis} for the details). 
For each  $\sigma(p_i)$, we can choose to apply either \lemref{lem:extractmv}~\cite{rose1958, mundici1994constructive}, which delivers an MV term of length $\Theta(2^{2^s})$,
or the optimized procedure \cite{aguzzoli1998complexity} 
which generates an MV term of length $\Theta((2^s)^2)$. The former choice results in overall length of at least $\Omega(2^{s+2^s})$, while the latter one yields at least $\Omega(8^{s})$.
In the simulation results below, we consistently apply the procedure in \cite{aguzzoli1998complexity}.

{\em Our algorithm with deep networks.}
We first note that $g$ can be realized by a ReLU network $\Phi_g$ according to $\Phi_g = \rho \circ W_2 \circ \rho \circ W_1 = g$ with
\[
W_1(x) = \begin{pmatrix}
    2 \\ 2 
\end{pmatrix} \! x \, - \begin{pmatrix}
    0 \\ 1 
\end{pmatrix}\hspace{-0.1cm}, \quad
W_2(x) = \begin{pmatrix}
    1 & -2 
\end{pmatrix} \begin{pmatrix}
    x_1 \\x_2 
\end{pmatrix}\hspace{-0.1cm}.
\]
Next, we run our algorithm described in Steps 1-3 in the proof of~\thmref{them:NN-MV} on $\Phi_g$. First, $\Phi_g$ is converted into the equivalent CReLU network $\Psi_g = \sigma \circ W_2^* \circ \sigma \circ W_1^* = \Phi_g$, with 
\[
W_1^*(x) = \begin{pmatrix}
    2 \\ 2
\end{pmatrix} x - \begin{pmatrix}
    0 \\ 1 
\end{pmatrix}\hspace{-0.1cm}, \quad
W_2^*(x) = \begin{pmatrix}
    1 &-1
\end{pmatrix} \begin{pmatrix}
    x_1 \\x_2 
\end{pmatrix}\hspace{-0.1cm}.
\]
Next, we apply Lemma~\ref{lem:extractmv} to the individual $\sigma$-neurons occurring in $\Psi_g$ to get
\begin{align} 
\label{eq:g11}\sigma(2x) & = x\oplus x \\
\label{eq:g12}\sigma(2x-1) & = x\odot x\\
\label{eq:g13}\sigma(x_1 -x_2) & = x_1\odot \lnot x_2.
\end{align} Finally, the overall MV term corresponding to the function $g$ is obtained by replacing $x_1$ in \eqref{eq:g13} by \eqref{eq:g11} and $x_2$ by \eqref{eq:g12} resulting in
\begin{equation}\label{eq:alg-g}
    \Psi_{g}(x)=(x\oplus x)\odot \lnot (x\odot x).
\end{equation}
We next establish that the self-compositions $g_s$ can be realized by ReLU networks and can hence be converted into equivalent CReLU networks. To this end, we first note that the ReLU network implementing $g_2$ is given by
\[
\Phi_g^2:=\Phi_g \circ \Phi_g =  \rho \circ W_2 \circ \rho \circ W_1 \circ \rho \circ W_2 \circ \rho \circ W_1,
\]
which has the equivalent CReLU network $\Psi_g^2:=\Psi_g \circ \Psi_g$. Likewise, the $s$-th order sawtooth function $g_s$, for $s\geq 3$, can be realized by CReLU networks according to
\[
\Psi^s_g: = \underbrace{\Psi_g \circ \ldots \circ \Psi_g}_{s}, \quad s \geq 3.  
\]
We can now conclude that the MV term corresponding to $\Psi^s_g$, for $s \geq 2$, is obtained from the right-hand-side of \eqref{eq:alg-g} by iteratively applying the following procedure $s-1$ times: Replace every occurrence of $x$ by the right-hand-side of \eqref{eq:alg-g}. The length of the resulting MV term is given by $4^s$.
Note that as opposed to the order-wise lower bounds above, here we compute the length of the extracted MV term precisely. 
We conclude that our algorithm, working off deep networks realizing $g_s$, provably produces MV terms of length comparable to or shorter than the other two methods. 

{\em Our algorithm with shallow networks.} Our algorithm offers flexibility the other two algorithms do not have, namely it can start from
different ReLU network realizations of a given McNaughton function. We now investigate the impact of this flexibility on the result produced by the algorithm. Specifically, we shall be interested in understanding how the network architecture, notably deep vs. shallow, influences the length of the MV term delivered.
This will be done by first converting 
the ReLU network $\Phi_g^s$ realizing $g_s$ into an equivalent shallow network $\phi_g^s$, given by 
\[
\phi_g^s = \rho \circ w_2 \circ \rho \circ w_1,
\]
with
\begin{align*}
     w_1(x) &=\begin{pmatrix}
        2^s \\ 2^s \\ 2^s \\ \vdots \\ 2^s
    \end{pmatrix}x+ \begin{pmatrix}
        0 \\-1 \\-2 \\ \vdots \\ -2^s+1
    \end{pmatrix}\hspace{-0.1cm}, \quad x\in \mathbb{R}, \\ 
     w_2(x) &= \begin{pmatrix}
        1 & -2 & 2 & -2 & 2 & \cdots & -2 
    \end{pmatrix} x, \quad x\in \mathbb{R}^{2^s},
\end{align*}
and then applying the algorithm. The corresponding CReLU network is given by 
\begin{equation}\label{eq:composition-gs}
  \psi_g^s = \sigma \circ w_2^* \circ \sigma \circ w_1^*,  
\end{equation}
where 
\begin{align*}
     w_1^*(x) &=\begin{pmatrix}
        2^s \\ 2^s \\ 2^s \\ \vdots \\ 2^s
    \end{pmatrix}x+ \begin{pmatrix}
        0 \\-1 \\-2 \\ \vdots \\ -(2^s-1)
    \end{pmatrix}\hspace{-0.1cm}, \quad x\in \mathbb{R}, \\ 
     w_2(x) &= \begin{pmatrix}
        1 & -1 & 1 & -1 &  \cdots & 1 & -1
    \end{pmatrix} x, \quad x\in \mathbb{R}^{2^s}.
\end{align*}
Application of Lemma~\ref{lem:length_sigma} now allows us to conclude that
each $\sigma$-neuron in the first layer of $\psi_g^s$ and the single $\sigma$-neuron in the second layer have associated MV terms of length $\Theta(2^{2^s})$. 
Composing MV terms according to \eqref{eq:composition-gs} hence shows that the MV term corresponding to $\psi_g^s$ has length
$\Theta(2^{2^{s+1}})$. This allows us to conclude that starting from a shallow network realization of $g_s$ leads to an MV term length that grows
double-exponentially in $s$, whereas working off a deep network realization as above yields length that is exponential
in $s$, specifically $4^s=2^{2s}$. This stark difference is also reflected in the simulation results below.
We finally note that 
ReLU networks realizing multi-dimensional (McNaughton) functions $f:[0,1]^n\rightarrow [0,1]$, with $n\geq 2$, do not always have an equivalent shallow counterpart \cite{haase2023lower}.

\subsection{Expressive power of deep networks}

\begin{figure}[tb]
\begin{center}
\centerline{\includegraphics[width=0.5\columnwidth]{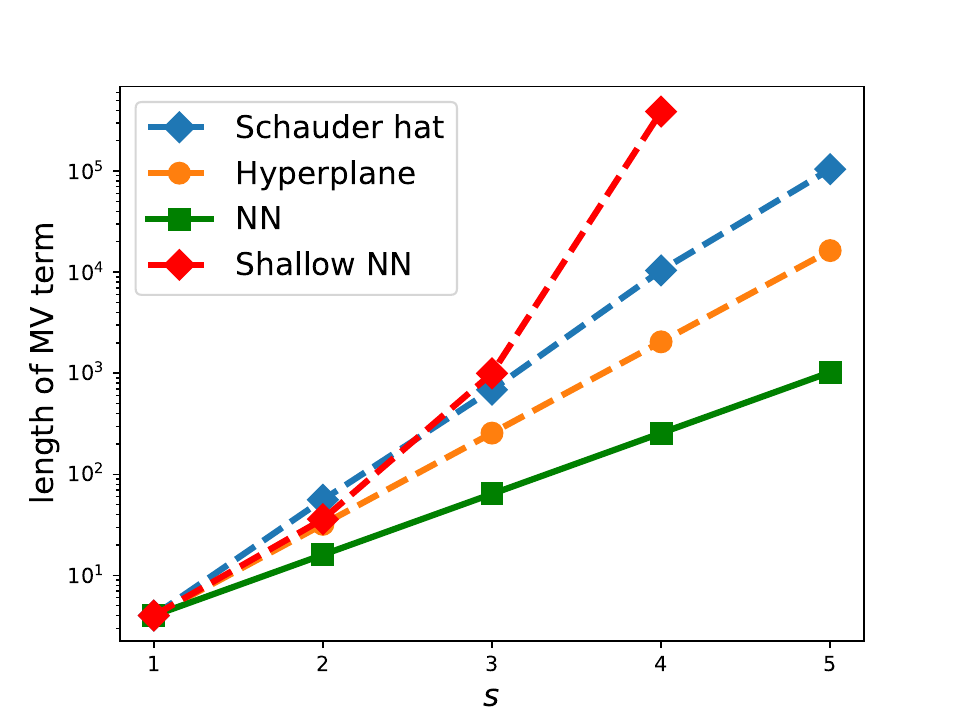}}
\caption{Length of the MV terms extracted from $g_s$ using different methods.}
\label{fig:res}
\end{center}
\vskip -0.2in
\end{figure}

The advantage in expressive power of deep networks over shallow ones has been analyzed in the neural network literature from various approximation-theoretic perspectives, see e.g. \cite{schmidt2020nonparametric, bianchini2014complexity, yarotsky2017error, raghu2017expressive, lin2017does, elbrachter2021deep, telgarsky2015representation}. 
The exponential vs. double-exponential dichotomy just identified establishes the representation benefits of deep networks from a mathematical logic perspective. 
In particular, our algorithm, when applied to the deep network representation $\Phi^{s}_g$ of the McNaughton function $g_s$, honors the compositional structure of $g_s = g  \circ  \ldots  \circ  g$ thereby rendering the algebraic expression of the resulting MV term compositional as well. Functional compositionality is hence turned into algebraic compositionality. In contrast, application of our algorithm to the shallow network representation $\phi^{s}_g$
of the McNaughton function $g_s$ ignores the compositional structure of $g_s$, which, in turn, leads to significantly longer MV terms. This observation
can be substantiated through simulation results.
Indeed, \figref{fig:res} 
displays double-exponential length growth behavior of our algorithm operating on the shallow network representation $\phi^{s}_g$ of $g_s$.
For the other three algorithms we see exponential length growth, as predicted by the analysis, with our algorithm working off a deep network resulting in
significantly shorter formulae than the Schauder hat and hyperplane methods.

The sawtooth function $g_s$ along with its associated MV term, obtained by iteratively self-composing \eqref{eq:alg-g}, constitutes an extreme case of compositional structure. We now investigate whether our algorithm retains an advantage over the Schauder hat and the hyperplane methods when the MV term underlying the McNaughton function does not exhibit compositional structure. To this end, we apply the three algorithms to 
McNaughton functions obtained from randomly generated MV terms of varying lengths without enforcing compositional structure\footnote{The specific details of the approach used to randomly generate MV terms along with code for reproducing all simulation results can be found at \url{https://www.mins.ee.ethz.ch/research/downloads/nn2mv_experiment.html}}. 
Specifically, for a randomly selected MV term $\tau(x)$ of length $L$, 
the corresponding (piecewise linear) McNaughton function $\tau^{\mathcal{I}}:[0,1]\rightarrow [0,1]$ is computed by 
exploiting the fact that its breakpoints in the interval $(0,1)$ 
are rational numbers
with denominators (in irreducible form) no larger than $L$ \cite{aguzzoli2006asymptotically, aguzzoli2000finiteness}.
It hence suffices to evaluate $\tau^{\mathcal{I}}(x)$ at the points $\{a/b:a,b\in \mathbb{Z}, 1\leq b\leq L\} \cap [0,1]$, which is done
by interpreting the connectives $\{\oplus, \odot,\lnot \}$ in $\tau(x)$ according to Definition 2.1. Each of the resulting samples of $\tau^{\mathcal{I}}(x)$ is then compared with its left and its right neighbor to determine whether it corresponds to a breakpoint. Finally, we perform linear interpolation between the breakpoints.
While the Schauder hat and the hyperplane methods are applied to the so-obtained McNaughton function $\tau^{\mathcal{I}}(x)$, 
our algorithm works off either the (deep) neural network obtained from $\tau(x)$ by following the constructive proof of 
the forward statement of \thmref{them:NN-MV} or the shallow (single-hidden-layer) network which identifies
the breakpoints of $\tau^{\mathcal{I}}(x)$ with individual neurons. 

\begin{figure}
\centering
\begin{subfigure}{.5\textwidth}
  \centering
  \includegraphics[width=.8\linewidth]{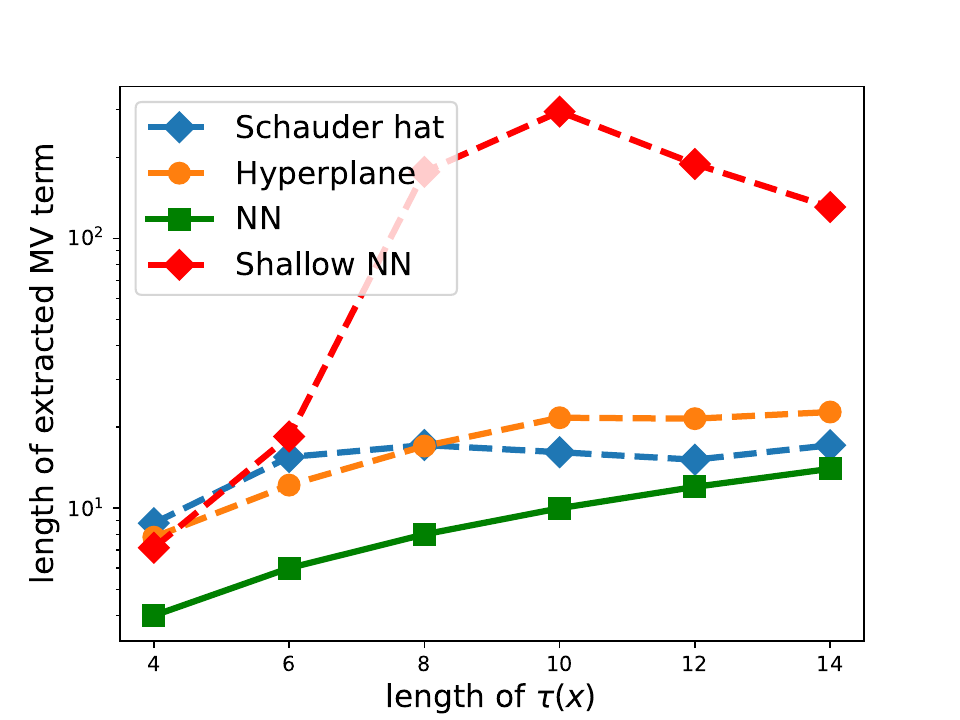}
  \label{fig:sub1}
\end{subfigure}%
\begin{subfigure}{.5\textwidth}
  \centering
  \includegraphics[width=.8\linewidth]{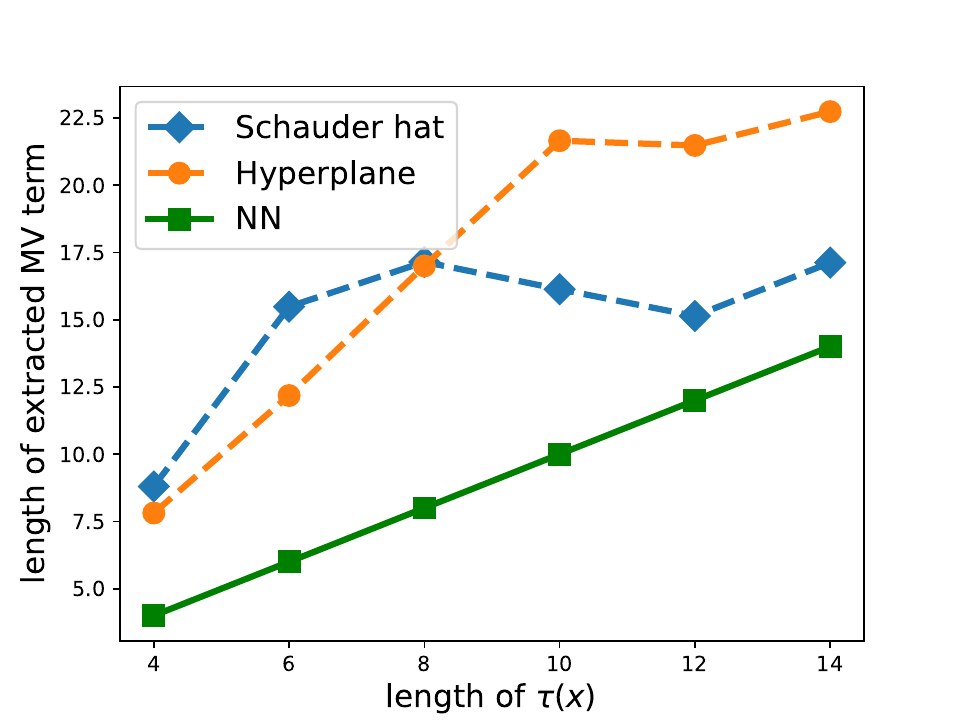}
  \label{fig:sub2}
\end{subfigure}
\caption{Lengths of MV terms extracted by different methods, the figure on the right depicts the same results as that on the left, but without the Shallow NN method.} 
\label{fig:sim-2}
\end{figure}

\begin{figure}[H]
\begin{center}
\centerline{\includegraphics[width=0.4\columnwidth]{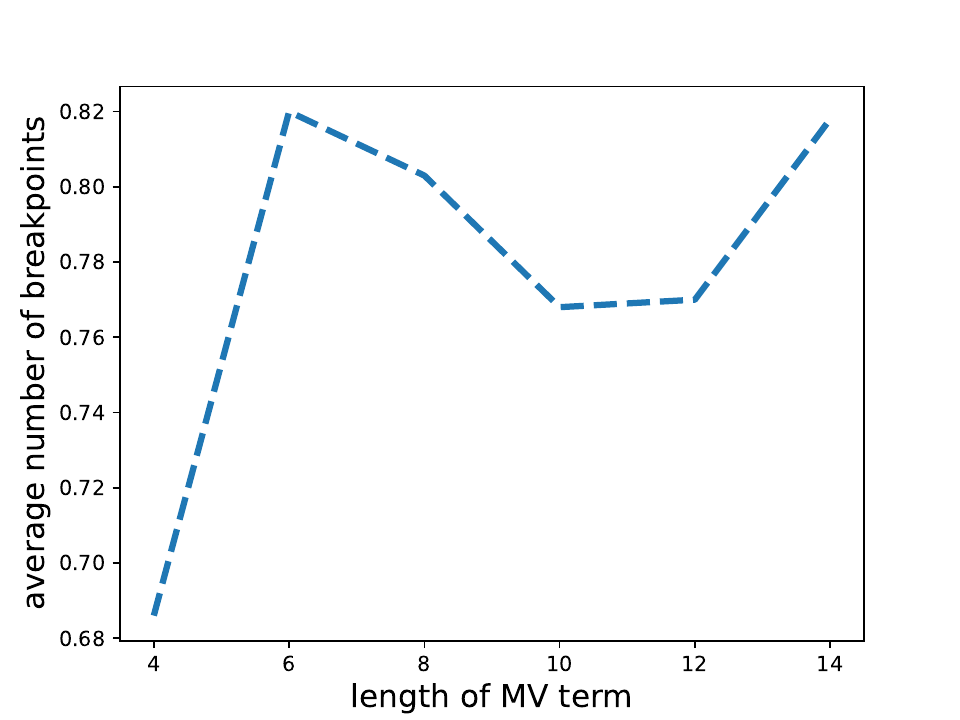}}
\end{center}
\vskip -0.5in
\caption{Average number of breakpoints of randomly generated MV terms.}\label{fig:0}
\end{figure}

For all methods, we verify that the MV terms delivered are, indeed, functionally equivalent to $\tau(x)$. This is done by comparing the breakpoints of
the associated McNaughton functions, obtained again through the sampling procedure described above. 
\figref{fig:sim-2} shows the average lengths of the extracted MV terms for $L$ ranging from $4$ to $14$ and 500 Monte Carlo runs for each value of $L$. 
We observe that our algorithm applied to deep network representations of $\tau(x)$ consistently achieves the shortest length among all four methods. In fact, it delivers MV terms whose lengths are almost identical to those of $\tau(x)$. This indicates that a deep network representation of the MV term $\tau(x)$ encodes algebraic information 
in a way not present in the shallow network representation. 
Our algorithm when applied to trained neural networks can hence honor algebraic structure present in the training data and expressed through the architecture of the network. 
The Schauder hat and the hyperplane methods do not have access to this
algebraic information as they operate on the functional representation of $\tau^{\mathcal{I}}(x)$. 
We finally note that the number of breakpoints of $\tau^{\mathcal{I}}(x)$ does not seem to consistently grow with the length of the underlying MV term $\tau(x)$.
\figref{fig:0} shows numerical results substantiating this claim, again for randomly generated MV terms and with $500$ Monte Carlo runs
for each data point. This observation also explains why
the extracted MV term lengths for the Schauder hat, hyerplane, and shallow NN methods in \figref{fig:sim-2} exhibit fluctuations as a function of $L$.
Concretely, we observed in our simulation studies that the lengths of the MV terms returned by these three methods are governed by
the number of breakpoints of the underlying McNaughton function.


\begin{figure}[H]
\begin{center}
\centerline{\includegraphics[width=0.5\columnwidth]{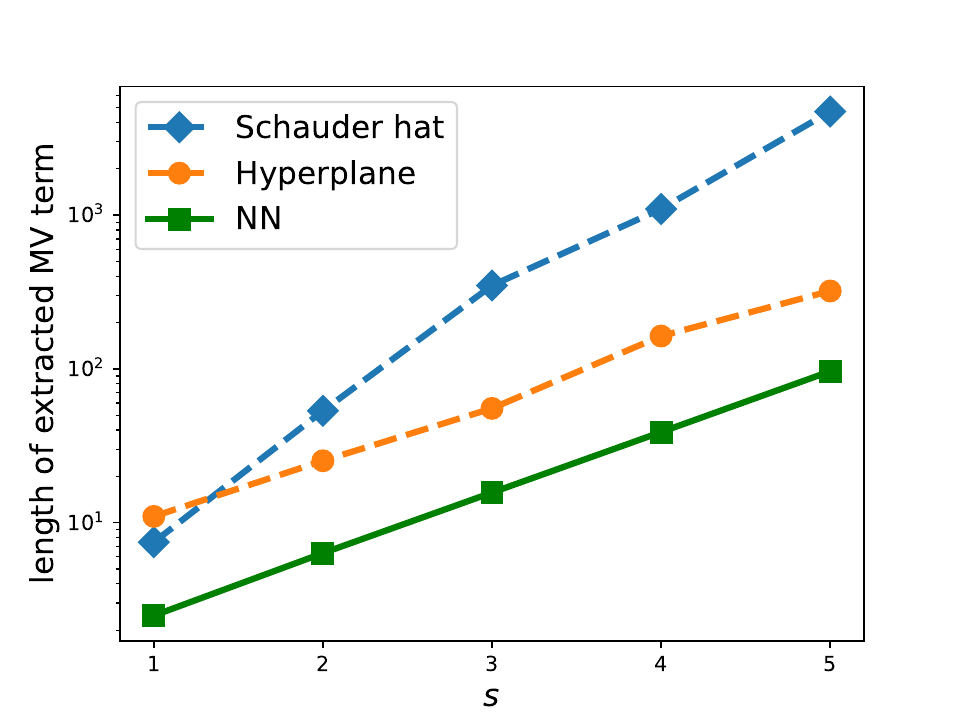}}
\end{center}
\vskip -0.5in
\caption{Lengths of MV terms extracted by different methods.}\label{fig:sim-3}
\end{figure}

We finally investigate whether our algorithm retains an advantage over the Schauder hat and the hyperplane methods 
when $\tau(x)$ exhibits general (as opposed to the extreme case corresponding to $g_{s}(x)$ analyzed above) compositional structure.
To see that this is, indeed, the case
we randomly generate $s$, for $s=1,\ldots,5$, MV terms $\tau_1(x),\ldots,\tau_s(x)$,
each of length $2$ or $3$, and then compute $\tau =\tau_1\circ \cdots \circ \tau_s$.
Again, all three methods deliver MV terms that are verified to be functionally equivalent to $\tau$, but have different lengths.
\figref{fig:sim-3} 
shows the results with 500 Monte Carlo runs for each value of $s$.
We can see that the MV term length increases exponentially in $s$ in all three cases, but again with 
the neural network algorithm resulting in significantly smaller lengths than the other two methods.

\subsection{The multi-dimensional case}

\begin{figure}[H]
\begin{center}
\centerline{\includegraphics[width=0.5\columnwidth]{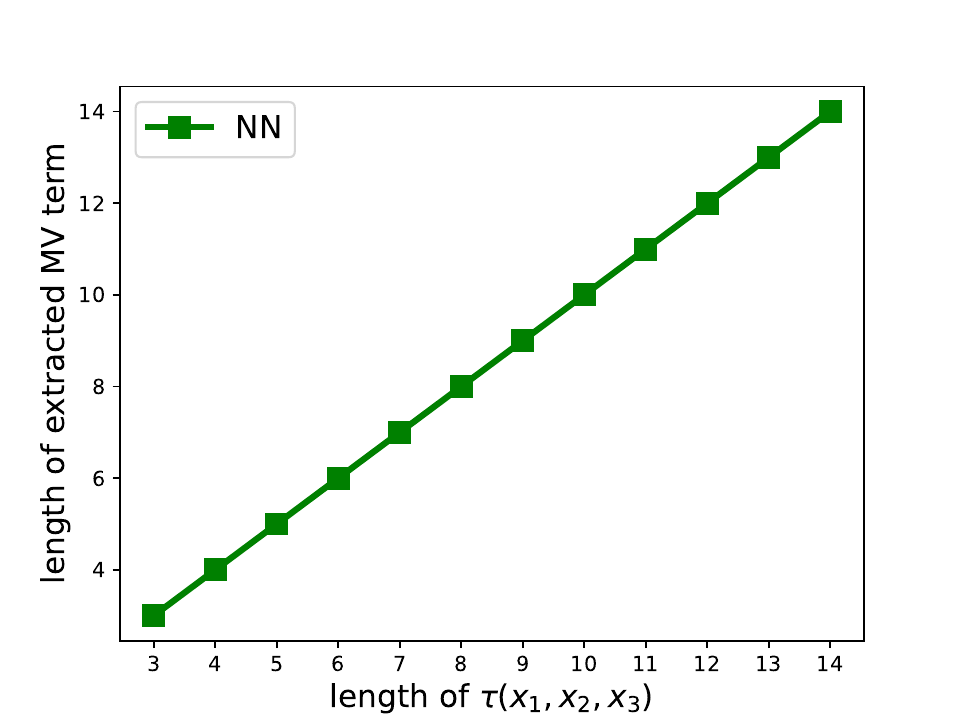}}
\end{center}
\vskip -0.5in
\caption{Length of MV terms extracted from three-dimensional McNaughton functions.}\label{fig:sim6}
\end{figure}

As mentioned in \secref{sec:review}, the Schauder hat and the hyperplane methods apply to the one-dimensional case only, whereas
the neural network algorithm can handle the general multi-dimensional case. To illustrate this, Fig.~\ref{fig:sim6} shows results
obtained by applying our algorithm to neural networks obtained from randomly generated MV terms $\tau(x_1,x_2,x_3)$ of different lengths\footnote{The length of an MV term in $d$ variables is defined as the total number of occurrences of propositional variables $x_1,...,x_d$.}, with $500$ Monte Carlo runs for each data point.
We observe that the algorithm returns MV terms of lengths almost identical to those of $\tau(x_1,x_2,x_3)$.

\section{Extensions to the Rational and Real Cases}\label{sec:4}
We now describe the extensions of our results to Rational \Luka logic and $\mathbb{R}\mathcal{L}$ logic, which, as mentioned in the introduction, have truth functions that are continuous piecewise linear, but with rational and, respectively, real coefficients. 

\subsection{The Rational Case}
Rational \Luka logic extends MV logic by adding a division (by integers) operation. The algebraic counterpart is given by the so-called divisible many-valued (DMV) algebras~\cite{gerla2001rational}. 

\begin{definition}\label{def:DMV01}
Consider the MV algebra $\mathcal{I}$ in~\defref{def:MV01}. Define the unary operations $ \delta_i x = \frac{1}{i}x, x\in[0,1], \text{ for all } i\in \mathbb{N}$. The structure  $\mathcal{I}_{\text{d}}= \langle [0,1], \oplus, \lnot, \{\delta_i\}_{i\in \mathbb{N}}, 0\rangle$ is a DMV algebra~\cite{gerla2001rational}.
\end{definition}

The class of term functions in $\mathcal{I}_{\text{d}}$ is given by the continuous piecewise linear functions à la Theorem~\ref{them:McNaughtonTheorem} but with rational coefficients~\cite{gerla2001rational, baaz1999interpolation}, hereafter referred to as 
rational McNaughton functions. We next extend~\thmref{them:NN-MV} to the rational case. 

\begin{theorem}\label{them:NN-DMV}
For $n\in \mathbb{N}$, let $\tau(x_1,\ldots, x_n)$ be a DMV term in $n$ variables and $\tau^{\mathcal{I}_{\text{d}}}:[0,1]^n \rightarrow [0,1]$ the associated term function in $\mathcal{I}_{\text{d}}$. There exists a ReLU network $\Phi$ with rational weights, satisfying 
\[
 \Phi(x_1,\ldots, x_n)=\tau^{\mathcal{I}_{\text{d}}} (x_1,\ldots,x_n), 
\]
for all $(x_1,\ldots,x_n) \in [0,1]^n$. Conversely, for every ReLU network $\Phi: [0,1]^n\rightarrow [0,1]$ with rational weights, there exists a DMV term $\tau(x_1,\ldots,x_n)$ whose associated term function in $\mathcal{I}_{\text{d}}$ satisfies
\[
\tau^{ \mathcal{I}_{\text{d}} } (x_1,\ldots,x_n) = \Phi(x_1,\ldots, x_n), 
\]
for all $(x_1,\ldots,x_n) \in [0,1]^n$.
\end{theorem}

\begin{IEEEproof}
We already know from the proof of Theorem~\ref{them:NN-MV} how to realize the operations $\oplus, \odot$, and $\lnot$ with ReLU networks. 
The division operation $\delta_i: x\rightarrow \frac{1}{i}x$, for $i\in \mathbb{N}$, by virtue of being an affine transformation, is trivially realized by a single-layer ReLU network. 
ReLU network realizations of formulae in Rational \Luka logic are obtained by concatenating ReLU networks implementing the operations $\oplus, \odot, \lnot$, and $\{\delta_i\}_{i\in \mathbb{N}}$. Inspection of the proof of the ReLU network composition~\lemref{lem:ReLUConcatenating} reveals that the resulting ReLU network has rational weights. 

Next, we extend the algorithm described in~\secref{sec:extract} to extract DMV terms from ReLU networks with rational weights. Steps~1 and~3 remain unaltered. 
We only remark that the additional weights introduced by the replacement procedure \eqref{eq:sigma-replacement} in Step 1 are all integer-valued and hence the resulting $\sigma$-network is guaranteed to have rational weights.
Step 2 needs to be mofidied as the $\sigma$-neurons are now of the form \begin{equation}\label{eq:sigmaneuronD}
   h= \sigma(m_1x_1+\cdots+m_nx_n+b),
\end{equation} with $m_1,\ldots,m_n, b\in \mathbb{Q}$, rendering~\lemref{lem:extractmv}, which requires $m_1,\ldots,m_n, b\in \mathbb{Z}$, inapplicable. We employ an idea from \cite{baaz1999interpolation} to transform a given $\sigma$-neuron with rational coefficients $m_1,\ldots,m_n, b$ into multiple $\sigma$-neurons with integer coefficients.
Concretely, let~$s\in \mathbb{N}$ be the least common multiple of the denominators of $m_1,\ldots,m_n,b$.
Recognizing that
\begin{equation}\label{eq:sh}
 s\sigma(x) = \sigma(sx)+\sigma(sx-1)+\cdots+\sigma(sx-(s-1)),   
\end{equation}
for $x\in \mathbb{R}$, and setting $h_i = \sigma(s(m_1x_1+\cdots+m_nx_n+b)-i)$, it follows that
$h = \sum_{i=0}^{s-1}\frac{1}{s}h_i$. As $h = \sum_{i=0}^{s-1}\frac{1}{s}h_i \leq 1$, the DMV term corresponding to $h$ is given by
$\oplus_{i=0}^{s-1}\delta_s\tau_i$, where $\tau_i$ denotes the DMV term associated with $h_i$.
\end{IEEEproof}

The Schauder hat and the hyperplane methods were extended to Rational \Luka logic in~\cite{gerla2001rational}, but again only for the univariate case.

\subsection{The Real Case}

We finally turn to the case of ReLU networks with real coefficients, which is of particular practical interest as it allows to extract logical formulae from trained ReLU networks. The Riesz many-valued algebra (RMV) \cite{di2014lukasiewicz} extends the MV algebra in~\defref{def:MV01} by adding a multiplication operation. 
\begin{definition}\label{def:DMV01}
Consider the MV algebra $\mathcal{I}$ in~\defref{def:MV01}. Define the unary operations $ \Delta_r x = r x, x\in[0,1], \text{ for all } r\in [0,1]$. The structure  $\mathcal{I}_{\text{r}}= \langle [0,1], \oplus, \lnot, \{\Delta_r\}_{r\in [0,1]}, 0\rangle$ is an RMV algebra~\cite{di2014lukasiewicz}.
\end{definition}

The term functions of the corresponding logic $\mathbb{R}\mathcal{L}$ are continuous piecewise linear functions with real coefficients~\cite{di2014lukasiewicz}. Noting that 
the multiplication operation $\Delta_r: x\rightarrow rx$ is an affine transformation, it follows that 
ReLU networks realizing RMV term functions 
are obtained in the same manner as in the integer and rational cases, namely by composing the ReLU networks realizing the individual operations appearing in the RMV term under consideration. Conversely, for every ReLU network $\Phi: [0,1]^n\rightarrow [0,1]$ with real weights, there exists an RMV term $\tau(x_1,\ldots,x_n)$ whose associated term function in $\mathcal{I}_{\text{r}}$ satisfies
\[
\tau^{ \mathcal{I}_{\text{r}} } (x_1,\ldots,x_n) = \Phi(x_1,\ldots, x_n), 
\]
for all $(x_1,\ldots,x_n) \in [0,1]^n$.

We now generalize our algorithm to extract RMV terms from ReLU networks with real weights. Again, Steps 1 and 3 in~\secref{sec:extract} remain unaltered. In Step 2, instead of \lemref{lem:extractmv}, we apply the following result. 

\begin{lemma}[\!\cite{di2014lukasiewicz}]\label{lem:extractmvR}
Consider the function $f(x_1,\ldots, x_n) = m_1x_1+\cdots+m_nx_n+b, (x_1,\ldots,x_n) \in [0,1]^n$, with $ m_1,\ldots \hspace{-0.02cm},m_n, b\in \mathbb{R}.$ For all $m\in (0,1]$ and $i\in \{1,\ldots,n\}$, with $f_{\circ}(x_1,\ldots, x_n) = m_1x_1+\cdots+m_{i-1}x_{i-1}+(m_i-m)x_i+m_{i+1}x_{i+1}+\cdots+m_nx_n+b$,
it holds that
\begin{equation}\label{eq:lemma4.5}
        \sigma(f) = (\sigma(f_{\circ}) \oplus (mx_i)) \odot \sigma(f_{\circ}+1).
\end{equation}     
\end{lemma}
For completeness, the proof of Lemma~\ref{lem:extractmvR} is provided in Appendix~\ref{app:proofLem4.5}.
We demonstrate the application of~\lemref{lem:extractmvR} through a simple example. Consider the $\sigma$-neuron $\sigma\hspace{-0.05cm}\left(\frac{1}{\sqrt{2}}x_1-2x_2\right)$. First, apply~\lemref{lem:extractmvR} with $m=\frac{1}{\sqrt{2}}$ and $i=1$ to get 
\begin{equation}\label{eq:e0}
   \sigma \hspace{-0.05cm} \left(\frac{1}{\sqrt{2}}x_1-2x_2 \right) = \left(\sigma(-2x_2) \oplus \left(\frac{1}{\sqrt{2}}x_1\right) \right) \odot \sigma(-2x_2+1).
\end{equation}
As $\sigma(-2x_2) = 0$ and $0\oplus x =x$,~\eqref{eq:e0} reduces to $(\frac{1}{\sqrt{2}}x_1) \odot \sigma(-2x_2+1)$. Next, using
$\sigma(x) = 1-\sigma(1-x), x\in \mathbb{R}$, we obtain $\sigma(-2x_2+1)=1-\sigma(2x_2)=\lnot (x_2\oplus x_2)$.
The overall RMV term associated with $\sigma\hspace{-0.05cm}\left(\frac{1}{\sqrt{2}}x_1-2x_2\right)$ is therefore given by $\Delta_{\frac{1}{\sqrt{2}}}x_1 \odot \lnot (x_2\oplus x_2)$. 

The hyperplane method was extended to the real case in~\cite{di2014lukasiewicz}, again only for the univariate case. An extension of the Schauder hat method 
to the real case does not seem to be available in the literature, but can readily be devised. 


\section*{Acknowledgment}
The authors are grateful to Prof. Olivia Caramello for drawing their attention to the McNaughton theorem and, more generally, to MV logic.

\appendices

\section{MV algebra}\label{app:MValge}

\begin{definition}[\!\! \cite{cignoli2013algebraic}]
\label{def:mv-algebra}
A many-valued algebra is a structure $\mathcal{A} = \langle A, \oplus, \lnot,0\rangle$ consisting of a nonempty set~$A$, a constant $0 \in A$, a binary operation  $\oplus$, and a unary operation $\lnot$ satisfying the following axioms:  
\begin{subequations}
\renewcommand{\theequation}{\theparentequation.\arabic{equation}}
\begin{align}
 \label{eq:associativity}    x \oplus (y \oplus z) &= (x \oplus y) \oplus z\\ 
  \label{commutivity}   x\oplus y  &= y\oplus x\\
 \label{eq:NeutralZero}    x \oplus 0  &=x\\
   \label{eq:NOTNOT}  \lnot \lnot x  &= x\\
 \label{eq:NeutralOne}    x \oplus \lnot 0  &= \lnot 0\\
  \label{eq:wedge operations}   \lnot (\lnot x \oplus y)\oplus y  &= \lnot (\lnot y \oplus x)\oplus x.
\end{align}
\end{subequations}
\end{definition}
An MV algebra $\langle A, \oplus, \lnot,0\rangle$ is said to be nontrivial iff $A$ contains more than one element. In each MV algebra we can define a constant $1$ and a binary operation $\odot$ as follows: \begin{align}
  \label{eq:1}    1 &:=\lnot 0  \\
  \label{eq:odot}    x \odot y&:= \lnot (\lnot x \oplus \lnot y).
\end{align} The ensuing identities are then direct consequences of~\defref{def:mv-algebra}:
\begin{subequations}
\renewcommand{\theequation}{\theparentequation.\arabic{equation}}
\begin{align}
  \label{eq:associativity for times} x \odot (y \odot z) &= (x \odot y) \odot z\\ 
    x\odot y  &= y\odot x\\
   x \odot 1  &=x\\     
  \label{eq:0 for times}   x \odot  0  &=  0.
\end{align}
\end{subequations}

We will frequently use the notions of MV terms and term functions formalized as follows.

\begin{definition}[\!\! \cite{cignoli2013algebraic}]
\label{def:mv-term}
Let $n\in \mathbb{N}$ and $S_n = \{(, ), 0,\lnot, \oplus, x_1,\ldots,$ $x_n\}$. An MV term in the variables $x_1, \ldots, x_n$ is a finite string over $S_n$ arising from a finite number of applications of the operations $\lnot$ and $\oplus$ as follows. The elements $0$ and $x_i$, for $i=1,\ldots,n$, considered as one-element strings, are MV terms.
\begin{enumerate}
  \item If the string $\tau$ is an MV term, then $\lnot \tau$ is also an MV term. 
  \item If the strings $\tau$ and $\gamma$ are MV terms, then $(\tau \oplus \gamma)$ is also an MV term.
\end{enumerate}
We write $\tau (x_1, \ldots, x_n)$ to emphasize that $\tau$ is an MV term in the variables $x_1, \ldots,x_n$.
\end{definition}

For instance, the following strings over $S_2 = \{(, ), 0,\lnot, \oplus, x_1,x_2\}$ are MV terms: 
\[
0, x_1, x_2,\lnot 0, \lnot x_2, (x_1\oplus \lnot x_2).
\]
We shall always omit the outermost pair of brackets for conciseness, e.g., we write $ x_1\oplus \lnot x_2$ instead of $(x_1\oplus \lnot x_2)$. Besides, for brevity we use the symbols $1$ and $\odot$ according to~\eqref{eq:1} and~\eqref{eq:odot}, respectively. 

MV terms are logical formulae of purely syntactic nature. To endow them with semantics, an MV algebra must be specified. The associated truth functions, a.k.a. term functions which we define presently, are then obtained by interpreting the operations $\oplus$ and $\lnot$ according to how they are specified in the MV algebra.

\begin{definition}[\!\! \cite{cignoli2013algebraic}]\label{def:term-function}
Let $\tau(x_1, \ldots, x_n)$  be an MV term in the variables $x_1, \ldots, x_n$. For the MV algebra $\mathcal{A} = \langle A, \oplus, \lnot,0\rangle$,
the term function $\tau^\mathcal{A}(a_1,\ldots,a_n): A^n \rightarrow A$ associated with $\tau$ in $\mathcal{A}$ is obtained by substituting $a_i$ for all occurrences of the variable $x_i$ and interpreting the symbol $0$ and the operations $\oplus$,$\lnot$ as the corresponding symbol $0$ and operations $\oplus$,$\lnot$ in $\mathcal{A}$.
\end{definition}

\section{ReLU networks}\label{app:relunn}

\begin{definition}[\!\!\cite{elbrachter2021deep}]
\label{def:ReLUNN}
Let $L \in \mathbb{N}$ and $N_0,N_1, \ldots, N_L \in \mathbb{N}$. A ReLU network is a map $\Phi: \mathbb{R}^{N_0} \rightarrow \mathbb{R}^{N_L}$ given by 
\[ \Phi= \begin{cases} 
      W_1, & L=1 \\
      W_2 \circ \rho \circ W_1, & L=2 \\
      W_L \circ \rho \circ W_{L-1} \circ \rho \circ \cdots \circ \rho  \circ W_1, & L \geq 3
   \end{cases} \text{,}
\] where, for $\ell \in \{ 1,2,\ldots \hspace{-0.02cm}, L\}$, $W_\ell: \mathbb{R}^{N_{\ell-1}}\rightarrow \mathbb{R}^{N_\ell}, W_\ell(x):= A_\ell x+b_\ell$ are affine transformations with weight matrices $A_\ell = \mathbb{R}^{N_\ell\times N_{\ell-1}}$ and bias vectors $b_\ell \in \mathbb{R}^{N_\ell}$, and the ReLU activation function $\rho: \mathbb{R}\rightarrow \mathbb{R},\rho\hspace{0.02cm}(x):=\max\{0,x\}$ acts component-wise. The number of layers of the network $\Phi$, denoted by $\mathcal{L}(\Phi)$, is given by $L$. We denote by $\mathcal{N}_{d, d'}$ the set of ReLU networks with input dimension $N_0=d$ and output dimension $N_L = d'$. 
\end{definition}

The next result formalizes properties of ReLU network compositions.

\begin{lemma}[\!\! \cite{elbrachter2021deep}]
\label{lem:ReLUConcatenating}
Let $d_1, d_2, d_3 \in \mathbb{N}$, $\Phi_1 \in \mathcal{N}_{d_1,d_2}$, and $\Phi_2 \in \mathcal{N}_{d_2,d_3}$. There exists a network $\Psi \in \mathcal{N}_{d_1,d_3}$ with $\mathcal{L}(\Psi) = \mathcal{L}(\Phi_1)+\mathcal{L}(\Phi_2)$, and satisfying \begin{equation*}
    \Psi(x) = (\Phi_2\circ \Phi_1)(x), \quad \text{ for all } x \in \mathbb{R}^{d_1}.
\end{equation*}
\end{lemma}
\begin{IEEEproof}
The proof is based on the identity $x = \rho(x) - \rho(-x)$. First, note that by~\defref{def:ReLUNN}, setting $L_1=\mathcal{L}(\Phi_1)$ and $L_2=\mathcal{L}(\Phi_2)$, we can write \begin{equation*}
    \Phi_1 = W_{L_1}^1 \circ \rho \circ  W_{L_1-1}^1 \circ \cdots \circ \rho  \circ  W_{1}^1 
\end{equation*} and \begin{equation*}
    \Phi_2 = W_{L_2}^2 \circ \rho \circ  W_{L_2-1}^2 \circ \cdots \circ \rho \circ  W_{1}^2,
\end{equation*} 
with the appropriate modifications when either $L_1$ or $L_2$ or both are equal to $1$ or $2$.
Next, let $N_{L_1-1}^1$ denote the input dimension of $W_{L_1}^1$ and 
$N_1^2$ the output dimension of $W_1^2$.
We  define the affine transformations~$\widetilde{W}_{L_1}^1: \mathbb{R}^{N_{L_1-1}^1} \rightarrow \mathbb{R}^{2d_2}$ and~$\widetilde{W}_{1}^2:\mathbb{R}^{2d_2} \rightarrow \mathbb{R}^{N_{1}^2}$ according to
\begin{align*}
 \widetilde{W}_{L_1}^1(x) :&= \begin{pmatrix} \mathbb{I}_{d_2} \\  -\mathbb{I}_{d_2} \end{pmatrix} W_{L_1}^1(x) \\
 \widetilde{W}_{1}^2(x) :&= W_1^2 \left( \begin{pmatrix} \mathbb{I}_{d_2} &  -\mathbb{I}_{d_2} \end{pmatrix} x \right).
\end{align*} The proof is completed upon noting that the network 
\begin{equation*}
    \Psi:=  W_{L_2}^2 \circ \cdots \circ W_2^2 \circ \rho \circ \widetilde{W}_{1}^2 \circ \rho \circ \widetilde{W}_{L_1}^1 \circ \rho \circ W_{L_1-1}^1 \circ \cdots \circ W_1^1
\end{equation*} satisfies the claimed properties. 
\end{IEEEproof}
\begin{lemma}\label{lem:NNminmax}
    There exist ReLU networks $\Phi^{\oplus}\in \mathcal{N}_{2,1}$ and $\Phi^{\odot } \in \mathcal{N}_{2,1}$ satisfying \begin{align*}
\label{eq:networkAND}    \Phi^{\oplus }(x,y) &= \min\{1,x+y\} \\
    \Phi^{\odot }(x,y) &= \max\{0,x+y-1\},
\end{align*} for all $x,y\in [0,1].$
\end{lemma}
\begin{IEEEproof}
First, to realize the operation $x\oplus y =\min\{1,x+y\} $, we note that addition can be implemented by a single-layer ReLU network according to
\begin{equation*}
    x+y =  \begin{pmatrix} 1 & 1 \end{pmatrix}  \begin{pmatrix} x \\y \end{pmatrix}.
\end{equation*}
It remains to implement the ``$\min$'' operation by a ReLU network. To this end, we observe that
\begin{equation*}
    \min\{1,x\} = 1-\rho(1-x) = (W_2 \circ \rho \circ W_1)(x),
\end{equation*} for $x\in [0,1]$, where
\begin{equation*}
    W_1(x) = -x+1, \quad \quad  W_2(x) = -x+1.
\end{equation*} 
Now, applying Lemma~\ref{lem:ReLUConcatenating} to concatenate the networks  $\Phi_1(x,y) = \begin{pmatrix} 1 & 1 \end{pmatrix}  \begin{pmatrix} x \\y \end{pmatrix}$ and $\Phi_2(x) = (W_2 \circ \rho \circ W_1) (x)$ yields the desired ReLU network realization of $x\oplus y$ according to
\begin{equation*}
    x \oplus y = (W_2^{\oplus} \circ \rho \circ W_1^{\oplus})(x,y),
\end{equation*} for $x,y\in [0,1]$, where
\begin{equation*}
   W_1^{\oplus}(x,y)= \begin{pmatrix}
       -1 & -1
   \end{pmatrix} \begin{pmatrix}
       x \\y
   \end{pmatrix}+1, \quad
   W_2^{\oplus}(x) = -x+1.
\end{equation*}

For the operation $x\odot y = \max\{0,x+y-1\}$, we simply note that \begin{equation*}
    \max\{0,x+y-1\} =  \rho \left( \begin{pmatrix}
        1 & 1
    \end{pmatrix}   \begin{pmatrix}
      x \\ y
    \end{pmatrix} -1 \right) = (W_2^{\odot} \circ \rho \circ W_1^{\odot})(x,y),
\end{equation*} for $x,y\in[0,1]$, where \begin{equation*}
    W_1^{\odot} (x,y) = \begin{pmatrix}
        1 & 1
    \end{pmatrix}   \begin{pmatrix}
      x \\ y
    \end{pmatrix} -1,\quad \quad  W_2^{\odot}(x) = x.
\end{equation*}
\end{IEEEproof}

\section{Proof of~\lemref{lem:extractmv}}\label{app:proofLem4.4}
  \begin{IEEEproof}
     We follow the line of arguments in~\cite{mundici1994constructive} and consider four different cases. \\
\noindent \textit{Case 1:} $f_{\circ}(x) \geq 1,$ for all $x\in [0,1]^n$. In this case, the LHS of~\eqref{eq:lemma4.4} is 
         \[
          \sigma(f)=1
         \]
and the RHS evaluates to
\begin{equation*}
      (\sigma(f_{\circ}) \oplus x_1) \odot \sigma(f_{\circ}+1)  =  (1 \oplus x_1)\odot 1 = 1.
\end{equation*}

\noindent \textit{Case 2:} $f_{\circ}(x) \leq -1,$ for all $x\in [0,1]^n$. In this case, the LHS of~\eqref{eq:lemma4.4} is 
     \[
          \sigma(f)=0
         \]
and the RHS satisfies
\begin{equation*}
      (\sigma(f_{\circ}) \oplus x_1) \odot \sigma(f_{\circ}+1) 
     =  (0 \oplus x_1)\odot 0=0. 
\end{equation*}

\noindent\textit{Case 3:}  $-1 < f_{\circ}(x)\leq 0$, for all $x\in [0,1]^n$. In this case, $f \in (-1,1]$ as $x_1\in [0,1]$. The RHS of~\eqref{eq:lemma4.4} becomes 
\begin{align*}
       (&\sigma(f_{\circ}) \oplus x_1) \odot \sigma(f_{\circ}+1) \\
     & =    (0 \oplus x_1)\odot (f_{\circ}+1) \\
     & =   x_1\odot (f_{\circ}+1)\\
    &  =  \max\{0,x_1+f_{\circ}+1-1\} \\
    &  =  \max\{0,f\}\\
    &  =  \sigma(f).
\end{align*}

\noindent \textit{Case 4:}  $0 < f_{\circ}(x) < 1$, for all $x\in [0,1]^n$. In this case, $f\in (0,2)$. The RHS of~\eqref{eq:lemma4.4} becomes 
\begin{align*}
     (& \sigma(f_{\circ}) \oplus x_1) \odot \sigma(f_{\circ}+1)\\
   &  =       (f_{\circ} \oplus x_1)\odot 1 \\
    & =f_{\circ} \oplus x_1\\
   &  =  \min\{1,f_{\circ}+x_1\}\\
   &  = \min \{1,f\}\\
   &  =  \sigma(f).
\end{align*}
 \end{IEEEproof}

\section{Proof of~\lemref{lem:extractmvR}}\label{app:proofLem4.5}
 We can follow the line of arguments in the proof of~\lemref{lem:extractmv} and consequently consider four different cases. \\
\noindent \textit{Case 1:} $f_{\circ}(x) \geq 1,$ for all $x\in [0,1]^n$. In this case, the LHS of~\eqref{eq:lemma4.5} is 
         \[
          \sigma(f)=1.
         \]
and the RHS evaluates to 
\begin{equation*}
      (\sigma(f_{\circ}) \oplus (mx_i)) \odot \sigma(f_{\circ}+1)  =  (1 \oplus (mx_i))\odot 1 = 1.
\end{equation*}

\noindent \textit{Case 2:} $f_{\circ}(x) \leq -1,$ for all $x\in [0,1]^n$. In this case, the LHS of~\eqref{eq:lemma4.5} is 
     \[
          \sigma(f)=0
         \]
and the RHS is given by 
\begin{equation*}
      (\sigma(f_{\circ}) \oplus (mx_i)) \odot \sigma(f_{\circ}+1) 
     =  (0 \oplus (mx_i))\odot 0=0. 
\end{equation*}

\noindent\textit{Case 3:}  $-1 < f_\circ(x)\leq 0$, for all $x\in [0,1]^n$. In this case, $f \in (-1,1]$ as $mx_i\in [0,1]$, for all $i \in \{1,\ldots,n\}$. The RHS of~\eqref{eq:lemma4.5} becomes 
\begin{align*}
      ( & \sigma(f_{\circ}) \oplus (mx_i)) \odot \sigma(f_{\circ}+1) \\
    &  =     (0 \oplus (mx_i))\odot (f_{\circ}+1) \\
   &   = (mx_i)\odot (f_{\circ}+1)\\
   &   =  \max\{0,mx_i+f_{\circ}+1-1\} \\
  &    = \max\{0,f\}\\
  &    =  \sigma(f).
\end{align*}

\noindent \textit{Case 4:}  $0 < f_{\circ}(x) < 1$, for all $x\in [0,1]^n$. In this case, $f\in (0,2)$. The RHS of~\eqref{eq:lemma4.5} becomes 
\begin{align*}
      (&\sigma(f_{\circ}) \oplus (mx_i)) \odot \sigma(f_{\circ}+1)\\
    & =      (f_\circ \oplus (mx_i))\odot 1 \\
    & =  f_\circ \oplus (mx_i)\\
   &  =  \min\{1,f_\circ+mx_i\}\\
  &   = \min \{1,f\}\\
  &   = \sigma(f).
\end{align*}

\section{Proof of Lemma \ref{lem:length_sigma}}\label{app:proof_length_sigma}
We prove the statement by induction on $m$. For the base case $m=1$, $\sigma(f)$ is of the form $\sigma(x_i+b)$ or $\sigma(-x_i+b)$, for some $i\in \{1,\ldots,n\}$ and $b\in \mathbb{Z}$. 
As $\sigma(x_i+b) \equiv 0$ for $b\leq -1$, $\sigma(x_i+b) \equiv 1$ for $b\geq 1$, $\sigma(-x_i+b) \equiv 0$ for $b\leq 0$, and $\sigma(-x_i+b) \equiv 1$ for $b\geq 2$, the only two cases where $\sigma(f)$ is not a constant are 
$\sigma(f)=\sigma(x_i)$, with corresponding MV term $x_i$, and $\sigma(f)=\sigma(1-x_i)$, with corresponding MV term $\lnot x_i$. 
Recalling that the length of an MV term is given by the total number of occurrences of propositional variables, this establishes the statement in the base case.
For the induction step, we conclude from \eqref{eq:lemma4.4} that the length of the MV term corresponding to $\sigma(f)$ equals the sum of the lengths of the MV terms corresponding to $\sigma(f_{\circ})$ and $\sigma(f_{\circ}+1)$ plus $1$. By the induction hypothesis, $\sigma(f_{\circ})$ has an MV term of length at most $2^{m-1}-1$. Further, the length of the MV term corresponding to $\sigma(f_{\circ}+1)$ is upper-bounded by that of the MV term associated with $\sigma(f_{\circ})$. We can therefore conclude that the length of the MV term corresponding to $\sigma(f)$ is no larger than $2^m-1$.

\section{Construction of the network $\Phi^\tau$ in  \secref{sec:extract}}\label{app:tau1}
By \lemref{lem:NNminmax}, we have the ReLU network associated with $x\oplus x$ according to
\[
W_2^\oplus \circ \rho \circ \left(\begin{pmatrix}
       -1 & -1
   \end{pmatrix} \begin{pmatrix}
       x \\x
   \end{pmatrix} +1 \right),
\] which can be rewritten as 
\begin{equation}\label{eq:xplusx}
    W_2^\oplus \circ \rho \circ  (-2x+1).
\end{equation}
As $y=\rho(y)-\rho(-y)$, for $y\in \mathbb{R}$, and $\lnot y = 1-y$, for $y \in [0,1]$, we obtain the $2$-layer ReLU network associated with $\lnot y$ according to
\begin{equation}\label{eq:noty}
   -\rho(y)+\rho(-y)+1.
\end{equation}
Finally, the ReLU network realizing the term function associated with $\tau=(x\oplus x)\odot \lnot y$ is obtained by composing the network $\Phi^\odot$ in \lemref{lem:NNminmax} with \eqref{eq:xplusx} and \eqref{eq:noty} according to
\[
W_2^\odot \circ \rho \circ W_1^\odot \circ \begin{pmatrix}
  W_2^\oplus \circ \rho \circ \left(-2x +1 \right)
    \\
  -\rho(y)+\rho(-y)+1
\end{pmatrix}\hspace{-0.1cm},
\]
which, through simple algebraic manipulations, is found to be equivalent to the network $\Phi^\tau$ in \eqref{eq:tau}.

\bibliographystyle{IEEEtran} 
\bibliography{IEEEexample}

\end{document}